\documentclass[10pt,twocolumn,letterpaper]{article}

\usepackage[pagenumbers]{iccv} %

\usepackage{booktabs}       %
\usepackage{amsfonts}       %
\usepackage{nicefrac}       %
\usepackage{microtype}      %
\usepackage{xcolor}         %
\usepackage{amsmath,mathtools}
\usepackage{amssymb}
\usepackage{amsthm}
\usepackage{graphicx}
\usepackage{enumitem}
\usepackage{bm}
\usepackage{bbm}
\usepackage{enumitem}
\usepackage{tabularx}
\usepackage{multirow}

\usepackage{tabularx}
\newcolumntype{L}{>{\raggedright\arraybackslash}X}
\newcolumntype{C}{>{\centering\arraybackslash}X}

\definecolor{mkcolor}{RGB}{255,0, 128}

\definecolor{donglaicolor}{RGB}{255, 87, 51}

\definecolor{nickcolor}{RGB}{0,0,255}

\definecolor{juncolor}{RGB}{0,0,255}

\definecolor{minghuacolor}{RGB}{0,128,128}

\def\eg{e.g.\ }               %
\def\ie{i.e.\ }               %

\newcommand{\parahead}[1]{\vspace{1mm}\noindent\textbf{#1.}\ }
\newcommand{\ourModel}{\textsc{PartField}\xspace}
\newcommand{\Shape}{S}

\newcommand{\ModelFunc}{f}
\newcommand{\PartProposal}{P}
\newcommand{\PointA}{\mathbf{p}_a}
\newcommand{\PointB}{\mathbf{p}_b}
\newcommand{\PointC}{\mathbf{p}_c}
\newcommand{\Point}{\mathbf{p}}

\makeatletter
\DeclareFontFamily{U}{MnSymbolC}{}
\DeclareFontShape{U}{MnSymbolC}{m}{n}{
    <-6>  MnSymbolC5
   <6-7>  MnSymbolC6
   <7-8>  MnSymbolC7
   <8-9>  MnSymbolC8
   <9-10> MnSymbolC9
  <10-12> MnSymbolC10
  <12->   MnSymbolC12}{}
\DeclareFontShape{U}{MnSymbolC}{b}{n}{
    <-6>  MnSymbolC-Bold5
   <6-7>  MnSymbolC-Bold6
   <7-8>  MnSymbolC-Bold7
   <8-9>  MnSymbolC-Bold8
   <9-10> MnSymbolC-Bold9
  <10-12> MnSymbolC-Bold10
  <12->   MnSymbolC-Bold12}{}
\DeclareSymbolFont{MnSyC}{U}{MnSymbolC}{m}{n}
\DeclareMathSymbol{\newcheckmark}{\mathord}{MnSyC}{160}
\makeatother
\usepackage{utfsym}
\newcommand{\newcrossmark}{\scalebox{0.75}{\usym{2613}}}

\definecolor{iccvblue}{rgb}{0.21,0.49,0.74}
\usepackage[pagebackref,breaklinks,colorlinks,allcolors=iccvblue]{hyperref}

\def\paperID{10578} %
\def\confName{ICCV}
\def\confYear{2025}

\title{
\ourModel: Learning 3D Feature Fields for Part Segmentation and Beyond
}

\author{Minghua Liu\thanks{Equal Contribution}~~\thanks{Work done during an internship at NVIDIA}~~$^{1,4}$ \;\;  Mikaela Angelina Uy\footnotemark[1]~~$^{1}$ \;\;  Donglai Xiang$^1$ \;\; Hao Su$^{4}$ \;\; 
\\ Sanja Fidler$^{1,2,3}$ \;\;  Nicholas Sharp$^1$ \;\;  Jun Gao$^{1,2,3}$ \\
 $^1$NVIDIA \quad $^2$University of Toronto \quad $^3$Vector Institute \quad $^4$UCSD
}

\begin{document}

\makeatletter
\let\@oldmaketitle\@maketitle
\renewcommand{\@maketitle}{\@oldmaketitle
    \centering
    \vspace*{-2em}
    \includegraphics{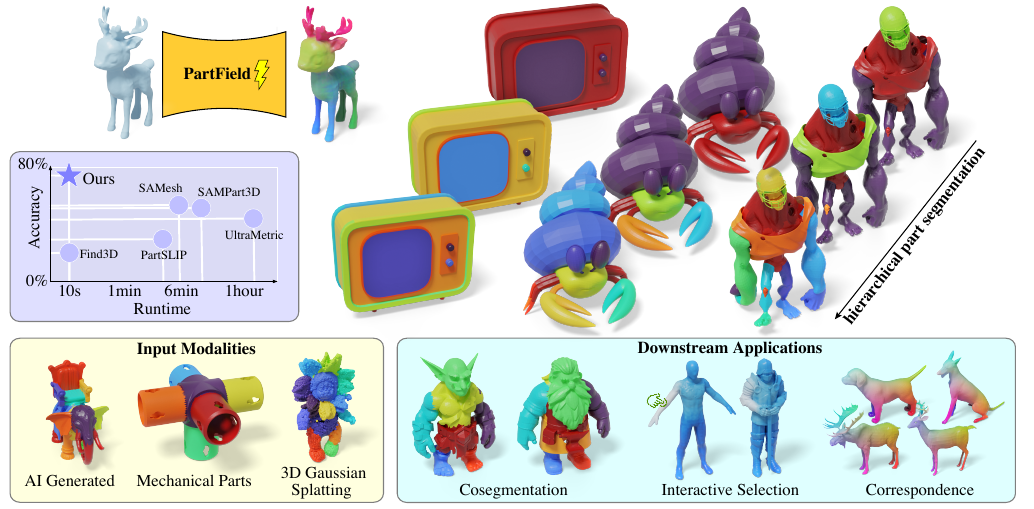}
    \vspace{-2em}
    \captionof{figure}{\footnotesize We propose \ourModel, a feedforward model that predicts part-based feature fields for 3D shapes. The learned features can be clustered to yield a high-quality part decomposition, and our method outperforms the latest open-world 3D part segmentation approaches in both quality and speed. \ourModel can be applied to a wide variety of inputs in terms of modality, semantic class, and style. The learned feature field exhibits consistency across shapes, enabling applications such as cosegmentation, interactive selection, and correspondence.}

    \label{fig:teaser}
\bigskip}
\makeatother

\maketitle
\begin{abstract}

We propose PartField, a feedforward approach for learning part-based 3D features, which captures the general concept of parts and their hierarchy without relying on predefined templates or text-based names, and can be applied to open-world 3D shapes across various modalities. PartField requires only a 3D feedforward pass at inference time, significantly improving runtime and robustness compared to prior approaches. Our model is trained by distilling 2D and 3D part proposals from a mix of labeled datasets and image segmentations on large unsupervised datasets, via a contrastive learning formulation. It produces a continuous feature field which can be clustered to yield a hierarchical part decomposition. Comparisons show that PartField is up to 20\% more accurate and often orders of magnitude faster than other recent class-agnostic part-segmentation methods. Beyond single-shape part decomposition, consistency in the learned field emerges across shapes, enabling tasks such as co-segmentation and correspondence, which we demonstrate in several applications of these general-purpose, hierarchical, and consistent 3D feature fields. Check our Webpage! \url{https://research.nvidia.com/labs/toronto-ai/partfield-release/}

\end{abstract}
    
\vspace{-1.2em}
\section{Introduction}
\label{sec:intro}
\vspace{-0.5em}

Human visual perception parses 3D shapes into parts based on their geometric structure, semantics, mobility, or functionality, while generalizing easily to new shapes. Such part-level understanding is critical to many applications, including 3D shape editing, physical simulation, robotic manipulation, and geometry processing. To empower machines with similar capabilities, 3D part segmentation has been studied as a standard task in computer vision~\cite{qi2017pointnet++,wang2019dynamic,thomas2019kpconv}, yet it remains challenging.

The first key challenge is data. Widely-adopted supervised learning-based methods~\cite{qian2022pointnext,jiang2020pointgroup,vu2022softgroup,qi2017pointnet++,wang2019dynamic,thomas2019kpconv} are limited by the lack of scale and diversity in 3D part-annotated datasets~\cite{mo2019partnet,Xiang_2020_SAPIEN}, leading to poor generalization to unseen categories of shapes. Following the success of large 2D image-based foundation models such as Segment Anything~\cite{sam,sam2,glip}, recent works have explored leveraging 2D priors to circumvent the reliance on 3D part annotations, and enabling open-world capability. 
However, most of these approaches~\cite{ultrametric,yang2024sampart3d,garfield2024} involve per-shape optimization. This requires a multi-step pipeline at inference time—rendering and segmenting multiple views, then fusing or distilling those segmentations into 3D—leading to lengthy runtimes (minutes to hours) and suffering from multi-view inconsistencies and noisy 2D predictions from 2D foundation models. In contrast, we aim to provide fast, direct predictions while generalizing across open-world shapes.

Another fundamental challenge is the definition of a ``part''. For example, at what granularity should parts be defined? Is a whole hand a part, or is each finger a separate part? Many past approaches~\cite{qi2017pointnet++,wang2019dynamic,thomas2019kpconv} including some recent open-world methods~\cite{liu2023partslip,find3d} attempt to match predefined part templates or text prompts.
However, committing to a predefined notion of parts impedes training at scale, as different data is inevitably inconsistent in part labels.
Text prompts come with their own complications: a single part might be referred to in multiple ways, and geometric parts may have no obvious language description. Instead, we aim to learn from large-scale data without committing to any single notion of parts. We want our model to cover a wide range of possible parts and multi-level decompositions, allowing applications to define the desired granularity and semantics.

This work proposes the \ourModel model for learning 3D parts and their hierarchy. Given a 3D shape as input, \ourModel predicts a continuous 3D feature field in a feedforward manner.
Instead of relying on part templates or text, the distance among \ourModel features implies the notion of parts: points with similar features are more likely to belong to the same part. The learned features can be queried continuously at any location, and can then be clustered to yield a part-aware, hierarchical decomposition of the shape, or even used for other downstream applications.

\begin{figure}
  \centering
  \includegraphics[width=\linewidth]{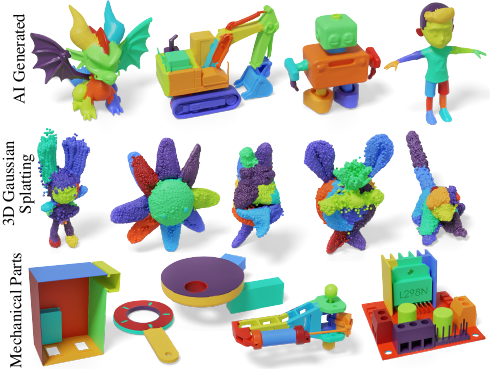}
  \vspace{-2em}
  \caption{\footnotesize \ourModel part segmentations across various 3D input modalities.
  \vspace{-1.5em}
  }
  \label{fig:multimodality}
\end{figure}

\ourModel is trained to match part proposals either predicted as image masks from existing 2D foundation models, or as explicit 3D supervision from existing datasets. There are no constraints on these part proposals; they can be defined based on semantic, geometric, or other criteria, and at any level of granularity. We leverage a carefully-chosen contrastive objective, encouraging samples from the same part to be more similar than samples for distinct parts ---this sidesteps the challenges of varying part granularity and differing notions of parts, enabling training on large-scale open-world data. By fitting a 3D-native feedforward model, \ourModel not only achieves dramatically faster inference, but also gains robustness to inconsistent and noisy labels. While we have not explicitly incorporated cross-shape supervision, we observe that our model emerges a feature space that is consistent across shapes—a useful property for downstream applications.

Our approach shares a similar philosophy—leveraging contrastive embedding learning and distilling image-space segmentations and features—with many recent or concurrent works on part segmentation \cite{ultrametric,garfield2024,yang2024sampart3d,ying2024omniseg3d, liu2024part123}, as well as numerous works on scene-level segmentation (see Section~\ref{sec:related_work} for a full list). However, we show that adapting and extending these techniques into a feedforward model trained at scale in an open-world setting has significant benefits in quality, speed, and utility.
We evaluate \ourModel against the latest baselines on the class-agnostic part segmentation task, showing performance improvements of more than 20\% while being an order of magnitude faster. As shown in Figure~\ref{fig:teaser}, this enables high-quality hierarchical part decomposition.
The resulting model can be applied across modalities including in-the-wild meshes, point clouds, and 3D Gaussian splats (Figure~\ref{fig:multimodality}).
The cross-shape consistency in the feature field enables usages beyond part segmentation, such as co-segmentation, selection, and correspondence. The key advantages of \ourModel are summarized as:

\begin{itemize} 
\item We train a feedforward, 3D-native model that enables fast inference, robust performance, and large-scale priors. %
\item \ourModel learns a versatile concept of 3D parts, applicable to 3D shapes across modalities and sources. 
\item We utilize triplet-based contrastive learning to sidestep inconsistencies in the notion of parts, enabling training at scale from diverse 2D and 3D data.
\item \ourModel features are consistent across shapes, yield promising cross-shape applications as a bonus, such as correspondence and co-segmentation.
\end{itemize}

\vspace{-2mm}
\section{Related Work}
\label{sec:related_work}
\vspace{-2mm}
Traditional data-driven part segmentation approaches~\cite{qian2022pointnext,jiang2020pointgroup,vu2022softgroup} typically predefine a part template, follow a supervised learning pipeline, and suffer from the limited scale and diversity of 3D part-annotated datasets~\cite{mo2019partnet,yi2016scalable}. Recently, open-world 3D segmentation has made significant progress, largely due to the success of 2D image foundation models~\cite{kirillov2023segment,ravi2024sam,li2022grounded,radford2021learning}. These methods can generally be divided into the following categories according to their tasks:

\noindent\textbf{Point-Prompt 3D Segmentation} Inspired by 2D SAM~\cite{sam}, some methods~\cite{zhou2024point,lang2024iseg} train feedforward models that take 3D points as prompts to segment 3D parts. SAM2POINT~\cite{guo2024sam2point} converts 3D input into videos and enables on-the-fly inference with SAM2~\cite{sam2}, which also supports point prompts. Our method addresses the more general problem of part-based feature learning and automatic hierarchical decomposition, and the learned field can also be applied to point-based interactive part selection.

\noindent\textbf{Text-Prompt 3D Segmentation} To enable text-based 3D segmentation, many works~\cite{abdelreheem2023satr,liu2023partslip,kim2024partstad} explore leveraging open-world 2D detectors by lifting and fusing predictions from multi-view rendering. Some works further explore combining both segmentation models and open-world detectors~\cite{abdelreheem2023zero,zhou2023partslip++,zhong2024meshsegmenter,xue2023zerops,huang2024openins3d,ton2024zero}. Another strategy involves aligning the 3D feature space with the CLIP~\cite{radford2021learning} text space, either through joint optimization with NeRF~\cite{kerr2023lerf} or Gaussian Splatting~\cite{qin2024langsplat}. Many works have also explored distilling text features into 3D networks for both scene segmentation~\cite{peng2023openscene,jiang2024open} and object segmentation~\cite{umam2024partdistill,find3d}, applying lightweight fine-tuning on top of CLIP~\cite{zhu2023pointclip,zhang2022pointclip}, or using CLIP feature for retrieval~\cite{takmaz2023openmask3d,lu2023ovir3dopenvocabulary3dinstance}. For 3D part segmentation, text prompts may not be ideal because many parts lack inherent semantic meaning. Instead, our work aims to learn from large-scale data what class-agnostic 3D parts could be, covering all possible parts—whether they are semantic or purely geometric.

\noindent\textbf{Class-Agnostic 3D Segmentation}  Many works leverage 2D SAM to achieve class-agnostic 3D segmentation. Instead of identifying and segmenting a single 3D part or object using a point prompt, they focus on segmenting or decomposing the entire input. For example, various methods have been developed to lift and merge multi-view 2D SAM prediction labels for both scene-level~\cite{yang2023sam3d,xu2023sampro3d,yin2024sai3d,guo2024sam,nguyen2024open3dis,xu2024embodiedsam,he2024pointseg} and object-level~\cite{tang2024segmentmeshzeroshotmesh} 3D segmentation. Some works attempt to train a feedforward scene segmentation model using SAM labels as pseudo-labels~\cite{huang2024segment3d} or by distilling SAM features~\cite{peng2023sam}, but mainly focus on scene level segmentation. Other approaches explore marrying SAM with NeRF~\cite{cen2023segment,ying2024omniseg3d,liu2024sanerf,gu2024egolifter,he2024view,chen2023interactive,garfield2024} or Gaussian Splatting~\cite{ye2024gaussian,cen2023segment,lan20242d,shen2024flashsplat,hu2024sagd,zhou2024feature} by adding an additional feature field to distill features or masks from 2D SAM. A recent work, Part123~\cite{liu2024part123}, integrates SAM into 3D reconstruction and enables part-aware single-image object generation. A line of traditional work also explores fine-grained 3D part segmentation models~\cite{luo2021learninggroupbottomupframework,wang2022learningfinegrainedsegmentation3d,yu2022partnetrecursivedecompositionnetwork}, but they are trained on closed-domain datasets and exhibit poor generalization to open-world scenarios. To the best of our knowledge, we are among the first to train a feedforward model that learns a feature field for open-world part decomposition.

\vspace{-2mm}
\section{Method}
\label{sec:method}
\vspace{-2mm}

\begin{figure*}[t]
    \centering
    \vspace{-3mm}
    \includegraphics{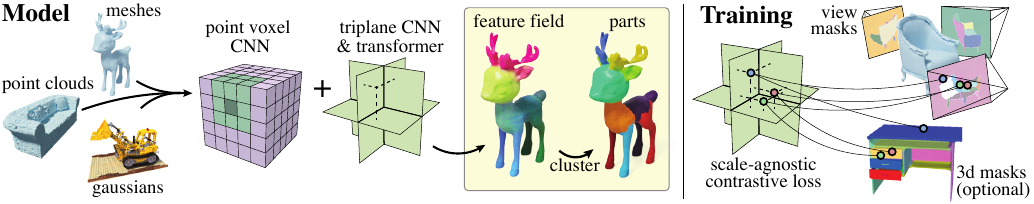}
    \vspace{-6mm}
    \caption{\footnotesize We train a feedforward model that takes a point-sampled 3D shape as input (which could come from a mesh, Gaussian splats, or other representations) and predicts a feature field represented by a triplane.
    These features can then be clustered to generate parts at various scales.
    Our model is trained with a contrastive loss on both open-world data, distilled from image-space masks, which need not be consistent, and 3D supervision when available.}
    \label{fig:model}
    \vspace{-3mm}
\end{figure*}

Given a 3D shape $\Shape$  as input, \ourModel predicts a continuous 3D feature field that encodes the underlying structure of parts and their hierarchy, in a feedforward manner.
As illustrated in Figure~\ref{fig:model}, this feature field maps any 3D point \(\Point\) to an $n$-dimensional latent feature vector \(\ModelFunc(\Point;\Shape): \mathbb{R}^3 \rightarrow \mathbb{R}^n\). The concept of parts is modeled by the feature distance between any two points: if two points \(\PointA\) and \(\PointB\) belong to the same part, their features \(\ModelFunc(\PointA;\Shape)\) and \(\ModelFunc(\PointB;\Shape)\) should be close in the latent feature space.

We first describe our data strategy, which involves collecting a diverse range of part proposals from multiple sources (Section~\ref{sec:part_proposal_and_point_triplets}). Next, we introduce our carefully chosen contrastive learning objective with hard negative mining, which captures multi-scale 3D parts and enhances training efficiency (Section~\ref{sec:contrastive_learning}). Finally, we present the architecture of our feedforward model (Section~\ref{sec:architecture}) and our inference strategy, which converts the learned part feature field into a hierarchical part decomposition (Section~\ref{sec:inference_and_applications}).

\vspace{-1mm}
\subsection{Training Part Proposals and Point Triplets}
\label{sec:part_proposal_and_point_triplets}
\vspace{-1mm}

\parahead{Part Proposals} We extract part proposals—which provide hints about which points should be grouped together to form a part—from both 2D and 3D data.
Each part proposal $P$ labels a subset of the shape $P \subset \Shape$ indicating that portion of the shape belongs to the same part. Note that we do not assume predefined part templates; therefore, the part proposals are not necessarily semantically associated. The proposals may come at any granularity and be defined by various criteria, such as geometry, semantics, or mobility.

Specifically, for 2D proposals, we follow previous work to render multi-view RGB and normal images of 3D shapes from large-scale unlabeled datasets~\cite{objaverse}. We then apply 2D foundation models (such as SAM2~\cite{sam2}) to predict class-agnostic 2D masks, which are subsequently projected back onto the shape. We densely sample point prompts, and each mask generates a proposal. Note that proposals from multiple masks are likely to overlap, covering various levels of granularity. For 3D proposals, we leverage part annotations available in existing 3D datasets~\cite{partnet}. Again, proposals may overlap if the labels have a hierarchical structure. 2D and 3D proposals complement each other: 2D proposals from image foundation models enable training on large unlabeled datasets and equip our model with open-world capabilities, while 3D proposals offer complete supervision of interior structure and capture valuable human semantic annotations.

\parahead{3D Point Triplets}
After obtaining part proposals, we sample triplets of 3D points from these proposals on the fly during training to apply a triplet-based contrastive loss. Specifically, for a given shape $\Shape$ and part proposal $\PartProposal$ on that shape, we sample a collection of triples of 3D points \(\{(\PointA, \PointB, \PointC)\}\) such that, within each triplet, points \(\PointA\) and \(\PointB\) come from the part proposal (i.e. \(\PointA, \PointB \in \PartProposal \subset \Shape\)), while \(\PointC\) comes from outside the part proposal (i.e. \(\PointC \in \Shape \setminus \PartProposal\)). For 2D proposals, all points in the triplets are sampled from the visible 2D pixels of the shape in that view, then unprojected into 3D using the known camera pose and object depth. Note that this means a 2D mask only contributes supervision on the visible surface of a shape. For 3D proposals, points are sampled directly from the labeled 3D geometry, including the interior space of the shapes and parts. When forming triplets, the positive points \(\PointA\) and \(\PointB\) are uniformly sampled, while the negative points are selected using a mining strategy discussed later. %

\vspace{-2mm}
\subsection{Contrastive Learning with Negative Sampling}
\label{sec:contrastive_learning}
\vspace{-1mm}

\begin{figure}[t]
    \centering
    \vspace{-3mm}
    \includegraphics[width=\linewidth]{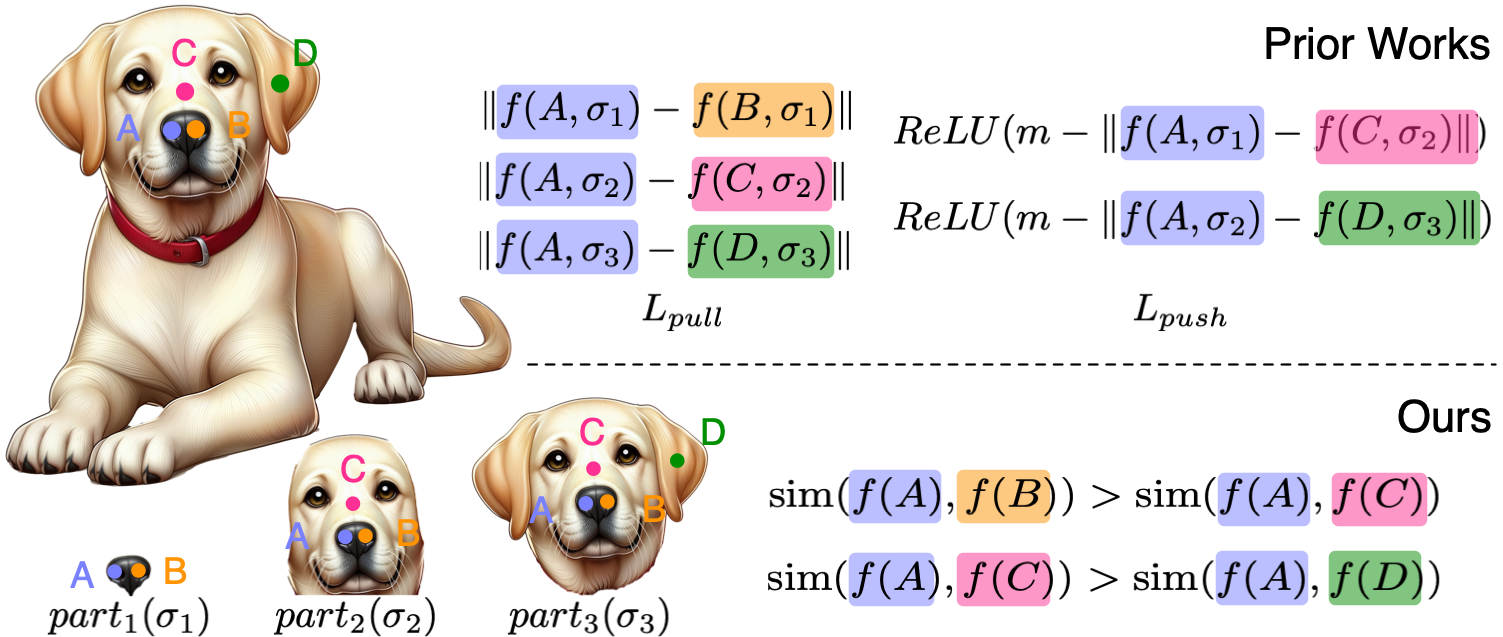}
    \vspace{-6mm}
    \caption{\footnotesize \textbf{(Left)} A point can belong to multiple parts at different scales.  
\textbf{(Upper Right)} Prior works~\cite{garfield2024,yang2024sampart3d} utilize pull and push losses to directly minimize or maximize the feature distances between point pairs, requiring an additional scaling condition to learn point features at different scales.  
\textbf{(Lower Right)} Our method employs a triplet loss that only encourages the relative relations between points within a triplet, enabling self-contained features ($\text{sim}(f(A),f(B)) > \text{sim}(f(A),f(C)) > \text{sim}(f(A),f(D)) $) that support multi-scale parts without need of scaling condition.} %
    \label{fig:loss}
    \vspace{-5mm}
\end{figure}

The essential idea of a contrastive triplet loss is to encourage the positive triplet points $\PointA,\PointB$ within a part proposal to be close, while the negative point $\PointC$ should be distant. However, the hierarchical and scale-ambiguous nature of part labeling complicates the design and scaling of such a loss. Figure~\ref{fig:loss} illustrates how two equally valid part proposals can assign points either to the same part or to different parts, depending on their levels of granularity. Prior work has attempted to impose an explicit scale conditioning parameter on the space~\cite{garfield2024,yang2024sampart3d}, but setting this scale consistently across data and supervision sources can be challenging.

Instead, we adopt a relative approach inspired by~\cite{chen2020simple}, which weakens the notion of supervision to simply encourage $\PointA$ to be closer to $\PointB$ than it is to $\PointC$, and likewise for $\PointB$.
The contrastive loss becomes
\vspace{-0em}
{
\scriptsize
\begin{align}
\label{eq:contrastive_loss}
\mathcal{L} = 
-\frac{1}{2}\Bigg(&\log\left(\frac{\text{sim}(\ModelFunc(\PointA),\ModelFunc(\PointB))}{\text{sim}(\ModelFunc(\PointA),\ModelFunc(\PointB))+ \text{sim}(\ModelFunc(\PointA),\ModelFunc(\PointC))}\right) + \notag \\
&\log\left(\frac{\text{sim}(\ModelFunc(\PointB),\ModelFunc(\PointA))}{\text{sim}(\ModelFunc(\PointB),\ModelFunc(\PointA))+\text{sim}(\ModelFunc(\PointB), \ModelFunc(\PointC))}\right)\Bigg)
\end{align}
}
where \(\text{sim}(\ModelFunc(\Point_u),\ModelFunc(\Point_v)) = \exp(\cos(\ModelFunc(\Point_u), \ModelFunc(\Point_v))/\tau\)) represents the exponential of the cosine similarity between two point features, and \(\tau\) is a learnable temperature. 
Related formulations have appeared in~\cite{ying2024omniseg3d,ultrametric,liu2024part123}.
Unlike directly minimizing the feature distance, our approach avoids conflicts when handling multi-scale proposals, sidesteps the need for an explicit scaling condition, and enables training on large datasets from many sources.

\begin{figure*}
    \centering
    \vspace{-3mm}
    \includegraphics{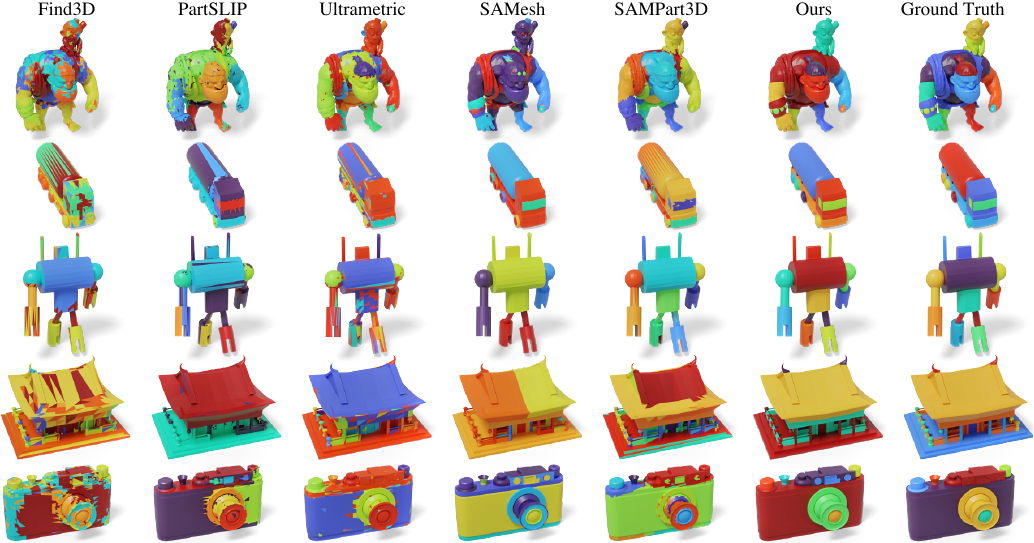}
    \vspace{-6mm}
    \caption{\footnotesize Qualitative comparison of class-agnostic segmentation on the PartObjaverse-Tiny dataset~\cite{yang2024sampart3d}. The baselines include Find3D~\cite{find3d}, PartSLIP~\cite{liu2023partslip}, Ultrametric Feature Field~\cite{ultrametric}, SAMesh~\cite{tang2024segmentmeshzeroshotmesh}, and SAMpart3D~\cite{yang2024sampart3d}. Each color represents a different part.} 
    \label{fig:main-com}
    \vspace{-5mm}
\end{figure*}

\parahead{Hard Negative Sampling} 
In Section~\ref{sec:part_proposal_and_point_triplets} we postponed the discussion of how to draw negative triplet samples $\PointC$ which do not belong to a particular part proposal.
A naive approach is to uniformly draw samples from complement of the part proposal, but we find training efficiency is improved by instead sampling more challenging negative points near part boundaries. Precisely, we sample negative points using a mix of three strategies, all drawn from the complement of the part proposal, \ie the unmasked visible surface in a 2D image view or the complement of a label in 3D.
\texttt{uniform} negatives are uniformly sampled, \texttt{3D-hard} prefers negatives closer to $\PointA$ in the Euclidean space, and \texttt{feature-hard} prefers negatives closer to $\PointA$ in feature space.
For efficiency, we evaluate the contrastive loss in parallel over many negative samples $\PointC$ for each positive sample $\PointA$ , summing over the $\textrm{sim}(\PointA,\PointC)$ term in the denominator of Equation~\ref{eq:contrastive_loss}.
The combination of these samples improves accuracy, especially near part boundaries (see Table~\ref{tab:ablation_part_objaverse_tiny} and Figure~\ref{fig:ablation_hard_negative}). Please refer to the supplementary material for details of the sampling.

\vspace{-1mm}
\subsection{Feedforward Model}
\label{sec:architecture}
\vspace{-1mm}

Unlike prior work~\cite{ultrametric,yang2024sampart3d,garfield2024,liu2023partslip} that utilizes per-shape optimization to lift or distill 2D predictions or priors, we instead train a feedforward 3D model $\ModelFunc(\Point,\Shape)$. 
This approach offers several benefits, including:  (a) fast inference;  (b) a consistent and complete 3D output feature field that smoothly extends to the shape interior;  (c) robustness against noisy and inconsistent part proposals, especially from 2D models; and (d) a unified feature space that naturally correlates across shapes, enabling additional downstream uses.

\parahead{Architecture}
Figure~\ref{fig:model} shows the architecture of our model. 
Input shapes, which may come as clean or in-the-wild meshes, point clouds, or even Gaussian particles~\cite{kerbl3Dgaussians} are sampled to a 3D point cloud.
This point cloud is used as input to the model, which outputs a feature field encoded as a triplane that can be evaluated at any spatial location.
The model consists of two main components.
First, a PVCNN~\cite{liu2019pvcnn} encoder extracts per-point features, which are then orthogonally projected via mean-reduction onto three axis-aligned 2D planes as an initial triplane representation.
These triplanes are then downsampled by a 2D CNN, flattened, passed to a transformer, then upsampled back to triplanes via a transposed 2D CNN.
Finally, for any 3D query, we retrieve and sum its corresponding features from the triplanes to evaluate the feature field for that point.

\begin{table*}[t]
  \centering
  \footnotesize
  \vspace{-3mm}
  \caption{\footnotesize Quantitative evaluation of class-agnostic part segmentation on PartObjaverse-Tiny~\cite{yang2024sampart3d} dataset. We use instance-level labels and report mean IoU.
  }
  \vspace{-3mm}
   \setlength{\tabcolsep}{2.7pt}
    \begin{tabular}{c|ccc|cccccccc|cc}
    \toprule
    \multirow{3}[4]{*}{Method} & \multicolumn{3}{c|}{Property} & \multicolumn{8}{c|}{Per Caetgory mIoU}                        & \multirow{3}[4]{*}{Average } & \multirow{3}[4]{*}{Runtime} \\
\cmidrule{2-12}          & feed- & multi-scale & text-input  & Human- & \multirow{2}[2]{*}{Animals} & Daily- & Buildings\& & Transpor- & \multirow{2}[2]{*}{Plants} & \multirow{2}[2]{*}{Food} & \multirow{2}[2]{*}{Electronics} &       &  \\
          & forward & feat field & free  & Shape &       & Used  & Outdoor & tations &       &       &       &       &  \\
    \midrule
    PartSLIP~\cite{liu2023partslip} & $\newcrossmark$     & $\newcrossmark$     & $\newcrossmark$     & 30.83 & 37.09 & 32.00 & 26.60 & 28.17 & 37.03 & 31.50 & 29.09 & 31.54 & $\sim$4min \\
    Find3D~\cite{find3d} & $\newcheckmark $     & $\newcrossmark$     & $\newcrossmark$     & 26.17 & 23.99 & 22.67 & 16.03 & 14.11 & 21.77 & 25.71 & 19.83 & 21.28 & \textbf{$\sim$10s} \\
    Ultrametric~\cite{ultrametric} & $\newcrossmark$     & $\newcheckmark $     & $\newcheckmark $     & 43.59 & 48.05 & 46.17 & 44.29 & 45.29 & 49.60 & 44.90 & 49.25 & 46.39 & $\sim$1.5h \\
    SAMesh~\cite{tang2024segmentmeshzeroshotmesh} & $\newcrossmark$     & $\newcrossmark$     & $\newcheckmark $     & 66.03 & 60.89 & 56.53 & 41.03 & 46.89 & 65.12 & 60.56 & 57.81 & 56.86 & $\sim$7min \\
    SAMPart3D~\cite{yang2024sampart3d} & $\newcrossmark$    & $\newcheckmark $     & $\newcheckmark $     & 55.03 & 57.98 & 49.17 & 40.36 & 47.38 & 62.14 & 64.59 & 51.15 & 53.47 & $\sim$15min \\
    Ours  & $\newcheckmark $     & $\newcheckmark $     & $\newcheckmark $     & \textbf{80.85} & \textbf{83.43} & \textbf{77.83} & \textbf{69.66} & \textbf{73.85} & \textbf{80.21} & \textbf{85.27} & \textbf{82.30} & \textbf{79.18} & \textbf{$\sim$10s} \\
    \bottomrule
    \end{tabular}%
  \label{tab:partobj}%
  \vspace{-3mm}
\end{table*}%

\vspace{-2mm}
\subsection{Inference and Clustering}
\label{sec:inference_and_applications}
\vspace{-1mm}
At inference, we apply the trained neural network once to generate the feature field triplanes, then sample part features as-desired, \eg at each face of a potentially high-resolution input mesh or even on the interior of a shape.
For mesh-based decomposition, we densely sample points from each face and use the average of these point features as the face feature. 
Although our network takes a 3D point cloud as input, we emphasize that it can also be applied to other 3D modalities (Figure~\ref{fig:multimodality}). For example, the 3D points may originate from 3D Gaussians or be sampled from a mesh surface. 

We can then apply off-the-shelf clustering algorithms to obtain a part-aware decomposition of the 3D shape.
Different settings may motivate different clustering strategies: $k$-means clustering is simple and fast, but agglomerative clustering yields crisp results on mesh connectivity.
Unless otherwise stated, we apply agglomerative clustering to mesh faces on the connectivity induced by face adjacency. The resulting hierarchical tree of parts can be leveraged in interactive applications, \eg in manual shape editing or rigging where users can adaptively select the branches that require further decomposition.

\vspace{-3mm}
\section{Experiments}
\label{sec:experiments}
\vspace{-1mm}

\begin{table}[t]
  \centering
  \scriptsize
  \caption{\footnotesize Quantitative evaluation of class-agnostic part segmentation on the PartNetE~\cite{liu2023partslip} test set. We use Instance-level labels and report mean IoU. Please refer to the supplementary material for the category group mapping. We do not report results from Ultrametric~\cite{ultrametric} because of the lengthy runtime on 1906 shapes.}
  \vspace{-3mm}
   \setlength{\tabcolsep}{0.8pt}
    \begin{tabular}{c|cccc|c}
    \toprule
          & PartSLIP~\cite{liu2023partslip} & Find3D~\cite{find3d} & SAMesh~\cite{tang2024segmentmeshzeroshotmesh} & SAMP3D~\cite{yang2024sampart3d} & Ours \\
    \midrule
    Electro. \& Comput.  & 20.79 & 14.58 & 23.96 & \textbf{43.86} & 43.70  \\
    Home Appliances &36.33 & 22.66 & 26.73 & 40.57 & \textbf{52.37}  \\
    Kitchen \& Food  & 40.78 & 24.89 & 31.50 & 65.18 & \textbf{69.85}  \\
    Furnit. \& Househo. &37.44 & 21.72 & 22.82 & 57.50 & \textbf{60.22} \\
    Tools, Offi., \& Misc. & 43.24 & 29.11 & 33.49 & 65.85 & \textbf{66.33}  \\
    \midrule
    average & 34.94 & 21.69 & 26.66 & 56.17 & \textbf{59.10}   \\
 
    \midrule
    runtime & $\sim$4min & \textbf{$\sim$10s } & $\sim$7min & $\sim$15min & \textbf{$\sim$10s} \\
    \bottomrule
    \end{tabular}%
  \label{tab:partnet}%
  \vspace{-3mm}
\end{table}%

\subsection{Implementation Details}
\vspace{-1mm}

\parahead{Training Datasets} We train our \ourModel on the Objaverse~\cite{objaverse} and PartNet~\cite{partnet} datasets. For Objaverse, we filter out low-quality data (e.g., LiDAR scans) and use the remaining approximately 340k shapes. For each shape, we render six RGB images and six mesh normal images, and feed each image into SAM2~\cite{sam2} for mask prediction. For each image, we densely sample a $32\times32$ grid of points as point prompts, generating 2D part proposals at various granularities. The PartNet dataset contains around 30k shapes with hierarchical part labels. We use part labels from all levels as our 3D part proposals. To sample 3D point triplets within each part, we convert the surface mesh from PartNet into a tetrahedral mesh using Tetgen~\cite{hang2015tetgen} and then take the vertices of tet mesh as sampled interior points.  All shapes are normalized to $[-1,1]^3$ during training, and we uniformly sample 100,000 points per shape as input to the network.

\parahead{Training Details} Our feature field is $448$-dimensional, the triplane spatial resolution is $512^2$, with 128 channels, and the transformer consists of 6 layers. Before feeding the triplane into the transformer model, we first downsample it to a resolution of 128 and treat each pixel as a token. After the transformer, we convert it back to $512^2$ triplanes. We train our model on 8 A100 GPUs for 2 weeks with a batch size of 2 per GPU. 

\vspace{-1mm}
\subsection{Comparison of Class-Agnostic Segmentation}
\vspace{-1mm}

\begin{figure*}
    \centering
    \vspace{-6mm}
    \includegraphics[width=\linewidth]{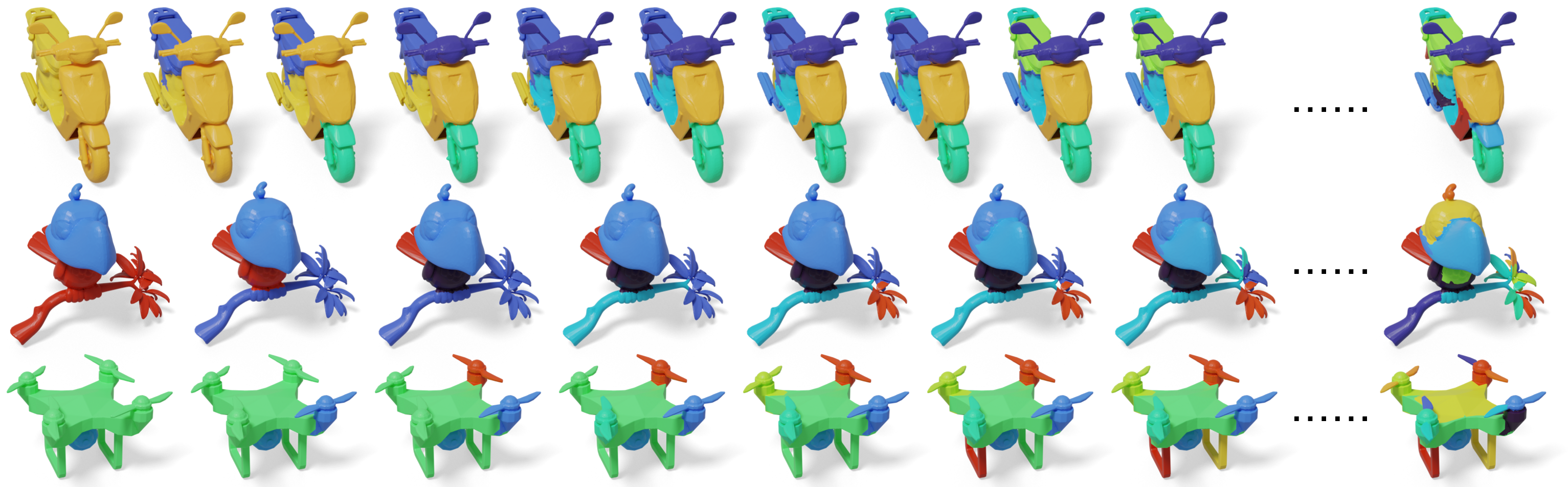}
    \vspace{-5mm}
    \caption{\footnotesize Hierarchical decomposition from \ourModel. Each row  from left to right reveals the structured decomposition at different levels of granularity. \ourModel  effectively captures meaningful hierarchical part relationships. Notice that in the first row, the handlebars and wheel are initially grouped into a single part with the front of the moped before splitting into individual parts.} 
    \label{fig:hierarchy}
    \vspace{-3mm}
\end{figure*}

Since \ourModel does not predict part semantics, we evaluate its performance on the class-agnostic segmentation task and compare to various latest baselines. 

\noindent\textbf{Evaluation Datasets}
We evaluate on PartObjaverse-Tiny~\cite{yang2024sampart3d} and PartNetE~\cite{liu2023partslip} datasets, following previous work on open-world 3D part segmentation~\cite{yang2024sampart3d,find3d,liu2023partslip}. The PartObjaverse-Tiny dataset contains 200 shapes spanning a wide range of object categories with human-annotated part segmentation. The PartNetE test set contains 1,906 shapes and is adopted from the PartNetMobility~\cite{Xiang_2020_SAPIEN} dataset, which covers 45 object categories with movable part annotations.

\noindent\textbf{Metric} We evaluate the class-agnostic mean Intersection over Union (mIoU) metric, following previous work~\cite{yang2024sampart3d,wang2021learning,xue2023zerops}. For each ground-truth part, we calculate the IoU with every predicted part and assign the maximum IoU as that part's IoU. We then compute the average IoU across all ground-truth parts. The IoU is computed between sets of mesh faces.

\noindent\textbf{Baselines} We compare \ourModel with five latest baselines for open-world 3D part segmentation, four of which are recent works that just appeared in 2024. As shown in Table~\ref{tab:partobj}, PartSLIP~\cite{liu2023partslip} and Find3D~\cite{find3d} are text-input part segmentation methods, whereas the other three baselines generate text-agnostic part segmentation. Except for Find3D—which trains a feedforward model to match the 3D feature space with the text feature space—all the other baseline methods employ per-shape optimization to lift or distill the results and features from 2D foundation models and require significant time to infer a single 3D shape. Among them, Ultrametric Feature Fields~\cite{ultrametric} performs NeRF optimization and takes multi-view images as input. For comparison, we first render multi-view images from our 3D input mesh to serve as the input of Ultrametric. While SAMPart3D~\cite{yang2024sampart3d} uses 3D pretraining to distill multi-view DINOv2~\cite{oquab2023dinov2} features into a 3D encoder, it still requires per-shape-based finetuning to distill from multi-view SAM predictions. SAMesh~\cite{tang2024segmentmeshzeroshotmesh} leverages a well-designed community detection algorithm to lift the multi-view predictions to 3D. 

For all methods, we follow their released codes evaluate on two datasets. For text-based methods, we use the part label names provided in the datasets. For methods that produce multi-scale part segmentations (i.e., Ultrametric, SAMPart3D, and Ours), we generate 20 segmentation results across different scales or cluster counts and select the one with the highest mIoU, following previous work~\cite{yang2024sampart3d}. When computing metrics, as not all approaches directly predict labels on input mesh faces, we carefully align and transfer the predicted labels to the input mesh faces for consistency. Please refer to the supplementary material for more details.

\noindent\textbf{Results}
Table~\ref{tab:partobj} presents the quantitative results on PartObjaverse-Tiny, where \ourModel significantly outperforms the baselines by a large margin (improving by $22.3\%$ over the second-best method) and operates several orders of magnitude faster. While most baselines require several minutes—or even over one hour (in the case of Ultrametric~\cite{ultrametric})—to process a single 3D shape, \ourModel can predict the feature field for an input 3D shape in a single feedforward pass, taking less than one second. We can then obtain the segmentation by applying the clustering algorithm to the features, which takes just a few seconds. Tables~\ref{tab:partnet} report the results on PartNetE, where a similar outperforming phenomenon is observed. We observe that, in general, text-based methods perform worse than class-agnostic approaches, especially in more open-world scenarios (e.g., PartObjaverse-Tiny). This indicates that accurately detecting 3D parts using open-world semantics remains a challenging task. Instead, \ourModel learn general concepts of 3D parts from large-scale data, enabling more accurate and diverse part decomposition.

Figure~\ref{fig:main-com} provides the qualitative comparison. In addition to the underperforming text-based methods, we observe that many per-shape optimization methods yield noisy results. This is mainly because the multi-view 2D predictions for a single shape can be inconsistent and imperfect, and the optimization process is sensitive to such noise. In contrast, \ourModel trains a feedforward model on large-scale noisy data, enabling it to predict more consistent and robust part segmentations. Note that while SAMesh~\cite{tang2024segmentmeshzeroshotmesh} produces decent results in many cases, it tends to generate overly fine-grained and sometimes over-segmented outputs (\ie the temple roof). Furthermore, SAMesh is neither flexible nor efficient in adjusting part granularity, whereas \ourModel efficiently supports flexible and adaptive multi-scale part decomposition.

\begin{figure}
  \centering
  \vspace{-2mm}
  \includegraphics[width=\linewidth]{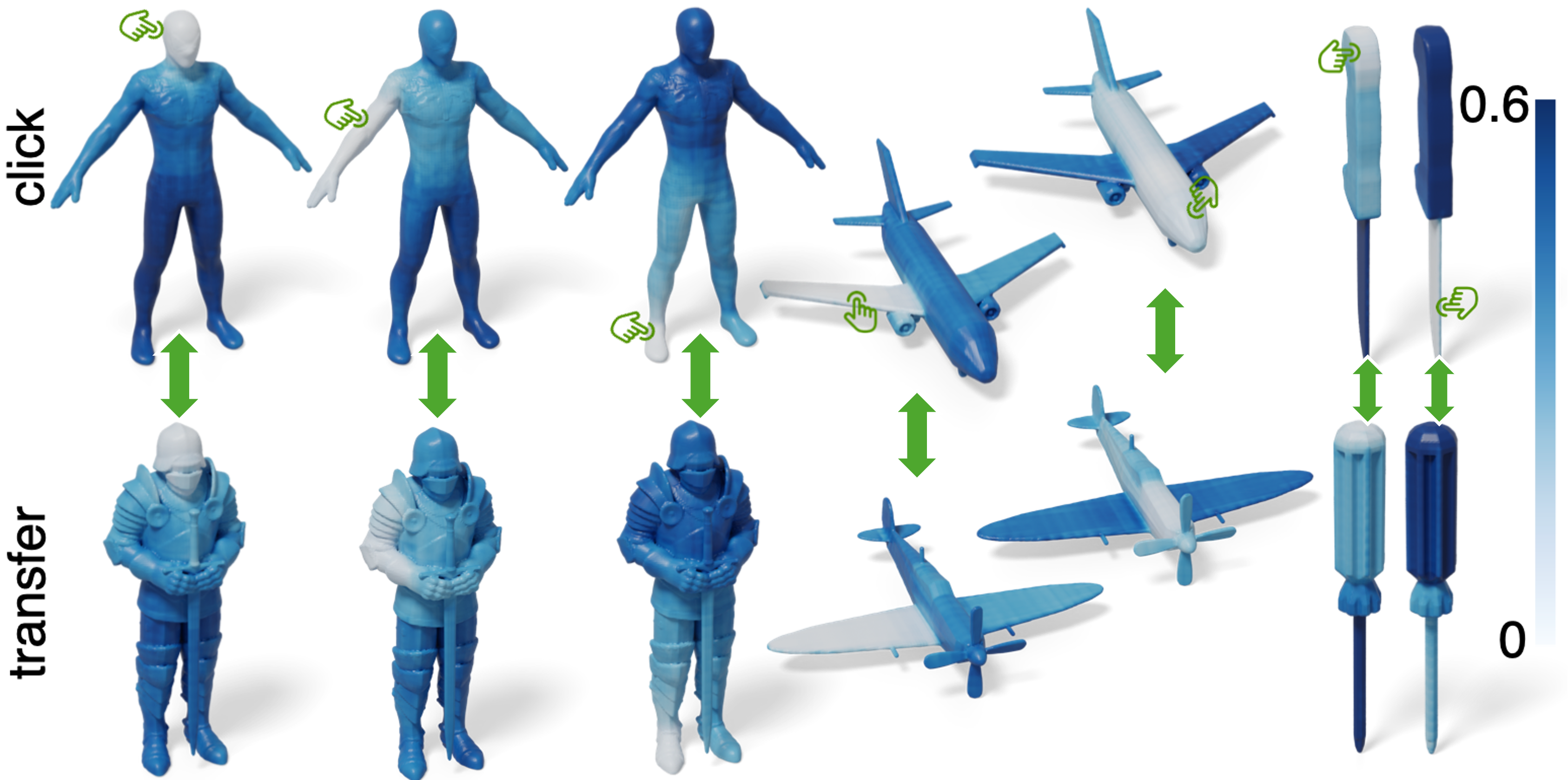}
  \vspace{-6mm}
  \caption{\footnotesize Feature Exploration. When a user clicks a point (top row), we display the similarity between that point and other regions within the same shape (top row), as well as similarities with locations on a different shape (bottom row).}
  \label{fig:feature_explore}
  \vspace{-4mm}
\end{figure}

\subsection{Applications}
\label{sec:applications}

We evaluate the properties of the learned feature field in various applications, including hierarchical part decomposition, 3D shape co-segmentation and 3D shape correspondences, feature field consistency. %

\noindent\textbf{Hierarchical Part Decomposition} \ourModel implicitly learns a hierarchy of multi-scale parts by contrastive learning at scale on a variety of 2D and 3D data.
A discrete hierarchical decomposition can be explicitly extracted via agglomerative clustering.
As shown in Figure~\ref{fig:hierarchy} (with additional results provided in the supplement), \ourModel effectively captures meaningful hierarchical relationships between parts, a capability which can be useful in many interactive applications. By exploring the hierarchical part tree, users can adaptively select which parts to decompose further.

\noindent\textbf{Feature Exploration} While we do not explicitly incorporate any cross-shape supervision, we find that consistency surprisingly emerges in our learned features space across different shapes. We develop an interactive interface to explore this consistency, visualizing similarity across the field to a selected location. As shown in Figure~\ref{fig:feature_explore}, semantic consistency emerges across shapes with different topologies (characters with different poses), geometries (airplanes), or functionalities (knife and screwdriver). This emerging property arises from our feedforward model design, which enables training our \ourModel at scale.%

\begin{figure}
  \centering
  \vspace{-2mm}
  \includegraphics{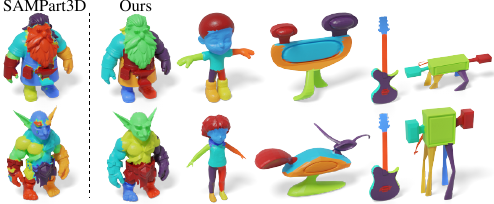}
   \vspace{-6mm}
  \caption{\footnotesize Quantitative results on co-segmentation. We co-segment the shapes from the top row with the corresponding one on the bottom row. The same color indicates the same part.}
   \vspace{-3mm}
  \label{fig:cosegmentation}
\end{figure}

\noindent\textbf{3D Shape Co-segmentation} We further explore the consistency of feature fields across shapes through a co-segmentation task. Specifically, we first segment the source shape to obtain the mean feature for each part. To segment the target shape, we use the mean feature as initialization and perform KMeans clustering to obtain its segmentation. As shown in Figure~\ref{fig:cosegmentation}, we successfully co-segment shapes with differing geometries and even establish correspondences on shapes with significant variations, such as an ogre and a bearded man. Furthermore, we compare our approach with SAMPart3D~\cite{yang2024sampart3d} using the same algorithm to obtain co-segmentation results. We observe that its performance is inferior, which may be attributed to its per-shape optimization training strategy.

\begin{figure}
  \centering
  \includegraphics{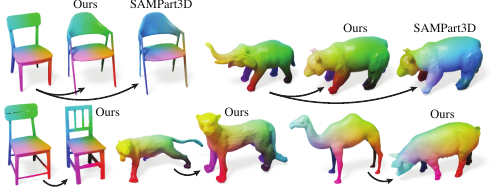}
  \vspace{-2em}
  \caption{\footnotesize Point-to-point correspondences obtained by Functional Maps~\cite{ovsjanikov2012functional} using our learned features as input. In each group, the colormap defined on the source shape (left) is transferred to the target shape (right). On the top row, we compare with the features from SAMPart3D.}
  \label{fig:correspondence}
\end{figure}

\noindent\textbf{3D Shape Correspondences} 
The cross-shape consistency from \ourModel allows it to serve as a large-scale prior for fine-grained point-to-point correspondence learning. We take Functional Maps \cite{ovsjanikov2012functional} for a promising initial example, fitting correspondence between a source and target shape.
We initialize correspondences between the shapes via nearest-neighbors in the \ourModel feature space, then refine these initial correspondences with Smooth Discrete Optimization \cite{magnet2022smooth}, iteratively solves for functional maps in a coarse-to-fine manner to recover a smooth point-to-point map. Figure~\ref{fig:correspondence} shows results, including a comparison applying the same strategy to SAMPart3D~\cite{yang2024sampart3d} features. The feature field from \ourModel provides accurate correspondence even if the topology of two shapes is very different (such as chairs on the bottom left), or with different poses (such as animals on the left).

\vspace{-0.2em}
\subsection{Ablations and Analysis}
\vspace{-0.2em}

\begin{figure}
    \centering
    \vspace{-2mm}
    \includegraphics[width=\linewidth]{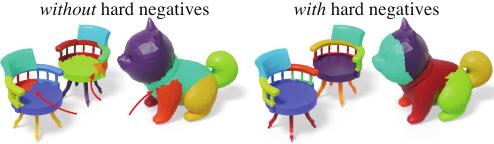}
    \vspace{-7mm}
    \caption{\footnotesize Qualitative results on ablating the hard negative sampling strategy. With hard negative mining, the part boundaries become much sharper.}
    \label{fig:ablation_hard_negative}
    \vspace{-4mm}
\end{figure}

\parahead{Hard Negative Mining and Training Triplet Source}
We conduct an ablation study on our hard negative training strategy and the source of training triplets using the PartObjaverseTiny dataset~\cite{yang2024sampart3d}. As shown in Table~\ref{tab:ablation_part_objaverse_tiny}, training on the large-scale, unlabeled 3D dataset Objaverse using only 2D part proposals already achieves quite decent results, demonstrating the effectiveness of this approach. Although additional 3D datasets are much smaller in scale (only 8\% of the Objaverse subset) and have limited diversity (only 24 categories), they still provide a modest gain in the open-world setting, suggesting their potential. Note that their benefits for supervising interior structures have not yet been reflected in this surface-only metric. Hard negative mining further improves performance and enables us to learn a crisper part boundary, as shown in Figure~\ref{fig:ablation_hard_negative} (see the chair's armrest). Incorporating both training triplets and the hard negative sampling strategy leads to the best overall performance.

\begin{table}[t]
  \centering
  \footnotesize
  
  \caption{\footnotesize Quantative results for ablation study. We report mIoU scores on PartObjaverse-Tiny dataset~\cite{yang2024sampart3d}.}
  \vspace{-2mm}
   \setlength{\tabcolsep}{2.3pt}
    \begin{tabular}{cc|cccc}
    \toprule
    \multicolumn{1}{c|}{\multirow{2}[2]{*}{Training Triplets}} & Objaverse (2D Proposals) & $\newcheckmark$     & $\newcheckmark$     & $\newcheckmark$     & $\newcheckmark$ \\
    \multicolumn{1}{c|}{} & PartNet (3D Proposals) &       & $\newcheckmark$     &       & $\newcheckmark$ \\
\cmidrule{1-2}    \multicolumn{2}{c|}{Hard Negative Sampling} &       &       & $\newcheckmark$     & $\newcheckmark$ \\
    \midrule
    \multicolumn{2}{c|}{mIoU} & 77.70 & 77.90 & 78.90 & 79.20 \\
    \bottomrule
    \end{tabular}%
  \label{tab:ablation_part_objaverse_tiny}%
\end{table}%

\parahead{Robustness to Input Modality} 
Thanks to the feed-forward model that takes point clouds as input, \ourModel can theoretically be applied to 3D shapes with other representations as well. To examine the model's robustness to various input modalities, data sources, and input styles, we evaluate it using diverse inputs, including AI-generated assets from both open-source models (Trellis~\cite{xiang2024structured}) and closed-source models (Edify3D~\cite{nvidia2024edify3d}, accessed via their public link), real-world 3D Gaussian splatting~\cite{kerbl3Dgaussians} (see the corresponding reconstruction results in the supplementary material), and CAD models of mechanical parts from the ABC dataset~\cite{Koch_2019_CVPR}. The results are shown in Figure~\ref{fig:multimodality}. While our method is trained mainly with point clouds from human-created meshes, we find it generalizes well to these various modalities, data sources, and input styles, suggesting that our method is widely generalizable and applicable.

\section{Limitations and Future Work}
\label{sec:conclusion}

Our PVCNN \& triplane architecture enables fast inference, but is inherently extrinsic, and as such our feature space is weakly correlated with 3D position---fortunately part segmentation is agnostic to this correlation, but cross-shape applications (Section~\ref{sec:applications}) do require the shapes to be consistently oriented.
Current investigations explore our \ourModel at the object-scale only, future work may extend to large scenes.
The cross-shape applications (Section~\ref{sec:applications}) are small-scale, investigating a surprising emergent property of the method---we hope that they motivate ongoing research into part contrastive learning for foundational features in 3D shape analysis.

\paragraph{Acknowledgements} We would like to additionally thank Masha Shugrina, Vismay Modi and team, for 3D scanned Gaussian splat assets and helpful discussions; and the Edify3D team, for the Edify assets and insightful discussions.

{
    \small
    \bibliographystyle{ieeenat_fullname}
    \bibliography{main}
}

\clearpage
\setcounter{page}{1}
\setcounter{section}{0}
\maketitlesupplementary
In this supplement, we provide details on our hard negative mining strategy, the baselines and additional qualitative results. We also refer readers to the included supplemental video, which gives additional results and renderings from multiple views.

\section{Details of Hard Negative Mining}
Positives and negatives for our contrastive learning training strategy are defined on each part mask (a SAM mask for 2D and a part label mask for 3D proposals). We first randomly sample $K$ masks for each batch of shapes. For each mask, we then sample $N=64$ positive pairs $(\PointA,\PointB)$ using the mask labels. Then we sample $M_1=256$ random negatives  $\PointC^1$ using \texttt{uniform}  strategy. For \texttt{3D-hard}  strategy, we sample $M_2=256$ hard negatives $\PointC^2$ also using the mask labels. The hard negatives are drawn based from a distribution weighted by the negative Euclidean distance of each negative candidate $\PointC$ to query $\PointA$. That is $\text{prob}(c_1^{(i)})=\frac{\text{dist}(c_1^{(i)}, A)}{\sum_c\text{dist}(c,A)}$. 
For \texttt{feature-hard}, we use a similar strategy as \texttt{3D-hard}, except the distance metric is computed on the feature space. 
Then the loss for a given triplet $(\PointA, \PointB, \{\PointC\})$ is as follows:

{\footnotesize
\begin{align}
        \mathcal{L}= 
    -\tfrac{1}{2}\Bigg(&\log\left(\frac{\text{sim}(\ModelFunc(\PointA),\ModelFunc(\PointB))}{\text{sim}(\ModelFunc(\PointA), \ModelFunc(\PointB))+\sum_{\PointC}\text{sim}(\ModelFunc(\PointA,\ModelFunc(\PointC))}\right) + \notag \\
    &\log\left(\frac{\text{sim}(\ModelFunc(\PointB),\ModelFunc(\PointA))}{\text{sim}(\ModelFunc(\PointB), \ModelFunc(\PointA))+\sum_{\PointC}\text{sim}(\ModelFunc(\PointB,\ModelFunc(\PointC))}\right)\Bigg)
\end{align}
}

\section{Details of Baseline Comparison}

\parahead{SAMPart3D~\cite{yang2024sampart3d}} 
We use the official codebase and the released checkpoint~\footnote{\url{https://github.com/Pointcept/SAMPart3D}}. The original code takes meshes as input and predicts mesh face labels. We directly uses the output to compute the metrics without any output alignment or label transfer. We obtain predictions across scales ranging from 0.0 to 2.0 at intervals of 0.125, compute the metric for each scale, and select the one with the best metric for each shape.

\parahead{Find3D~\cite{find3d}}  
We use the official codebase and the released checkpoint~\footnote{\url{https://github.com/ziqi-ma/Find3D}}. Following the instructions, we sample 5,000 points per shape, assigning them a dummy white color. We input the ground truth text prompts into the network, which then predicts point-wise labels corresponding to the text prompt. The point-wise labels are transferred to the input mesh by finding the nearest neighbor for the center of each input mesh face.  

\parahead{SAMesh~\cite{tang2024segmentmeshzeroshotmesh}}  
We use the official codebase~\footnote{\url{https://github.com/gtangg12/samesh}}. The original code takes meshes as input and predicts mesh face labels. For the PartObjaverseTiny dataset~\cite{xu2023sampro3d}, we use the default settings, which produce reasonably good results. However, the default settings fail for the PartNetE dataset~\cite{liu2023partslip}. We contacted the authors via email and confirmed that this issue arises due to the low-poly characteristics and poor mesh topology of the PartNet meshes. For the PartNetE dataset, we refined the hyperparameters and made modifications to the code following the authors' suggestions, achieving better results compared to the default setup.

\parahead{PartSLIP~\cite{liu2023partslip}} We used the official codebase and released checkpoint~\footnote{\url{https://drive.google.com/drive/u/0/folders/19j6PZfW8TDQ1ifHZwHIhn6X4BHjYRFCL}}. We employed the zero-shot version and utilized the instance segmentation predictions. Since the model takes a dense point cloud as input, we first rendered multi-view RGB images and depth maps of the input 3D meshes, which we then fused to obtain a colored dense point cloud. We fed both the dense point cloud and the ground truth part names into the network, which predicted the corresponding point-wise labels. These labels were transferred to the input mesh by finding the nearest neighbor to the center of each mesh face.

\parahead{Ultrametric Feature Field~\cite{ultrametric}} We used the official codebase~\footnote{\url{https://github.com/hardyho/ultrametric_feature_fields}}. Since the method takes multi-view images as input and performs a NeRF optimization, we first rendered multi-view images following the camera poses provided for the example asset in the code repository. We then followed the code instructions to generate multi-view SAM predictions and their corresponding hierarchy, and ran the NeRF optimization using the provided default configuration. After the NeRF optimization, we used ground truth multi-view depth to extract a fused point cloud and query its corresponding predicted features. Finally, we utilized the provided clustering algorithm, converted the clustering labels back to the 3D input mesh by finding the nearest neighbor for the center of each input mesh face, and evaluated 20 scales—computing the metric for each scale and selecting the one with the best performance for each shape.

\section{Multiclass Regression Interactive Cosegmentation}

In addition to the clustering-based cosegmentation described in Section 4.3, the supplemental video includes an additional demonstration where we interactively cosegment all of the \texttt{guitar} shapes from the COSEG dataset~\cite{wang2012active}.
The user clicks a small number of points on any shape, which are then extended in real-time to segmentations of all shapes, thanks to our precomputed feature field on each shape.

This also demonstrates an alternate learning-based setup using our feature field. Rather than clustering, we use user-annotated points as a (very small) training set to fit a logistic regression model for one-vs-rest multiclass classification, mapping from our features to the segmentation label. These models are easily fit and evaluated on all shapes in real time on the GPU, providing the user with helpful interactive feedback.

\begin{figure*}
  \centering
  \includegraphics[width=0.9\linewidth]{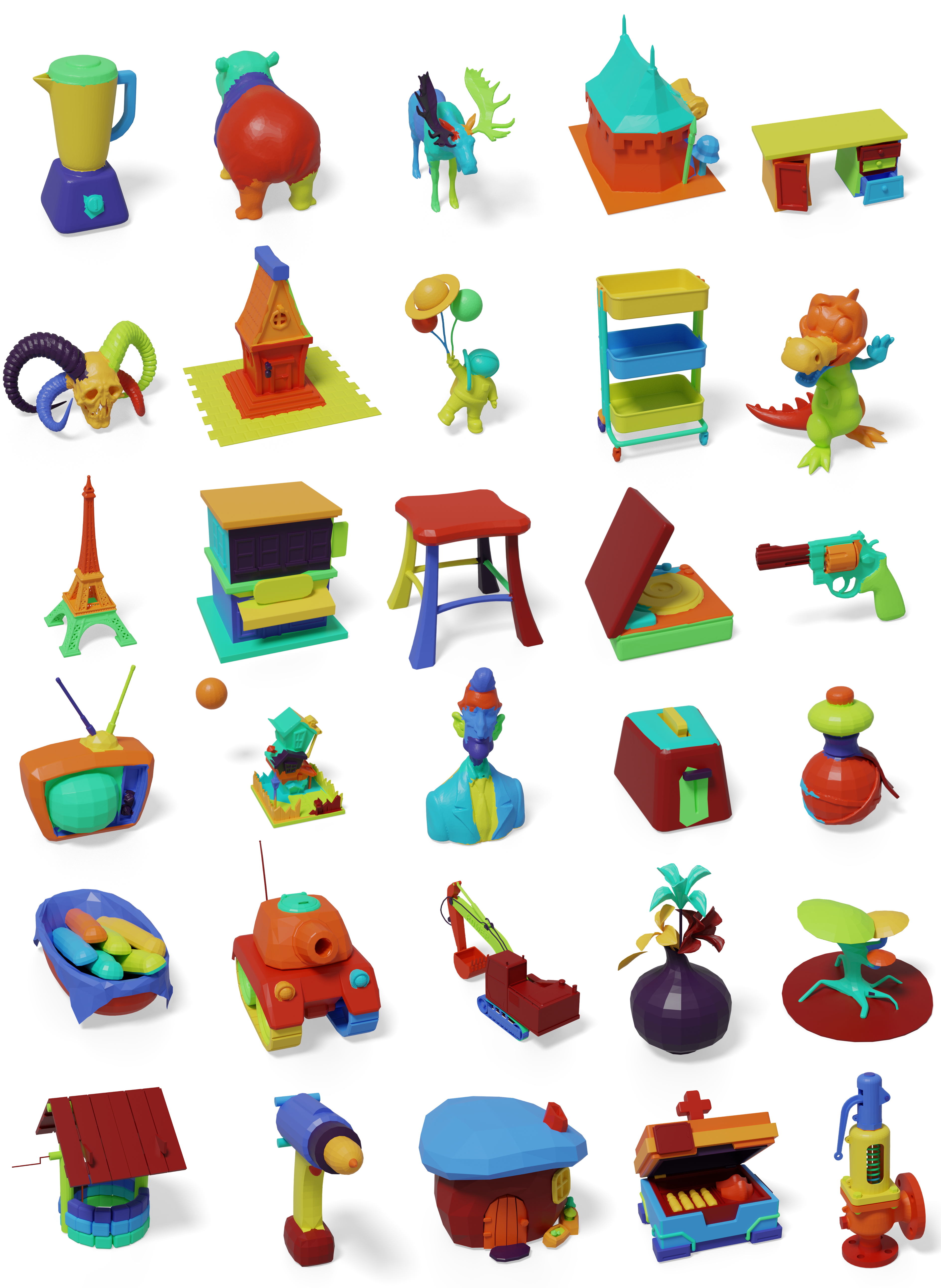}
  \caption{Additional qualitative examples segmentation results.}
  \label{fig:segmentation_supp}
\end{figure*}

\begin{figure*}
  \centering
  \includegraphics[width=0.75\linewidth]{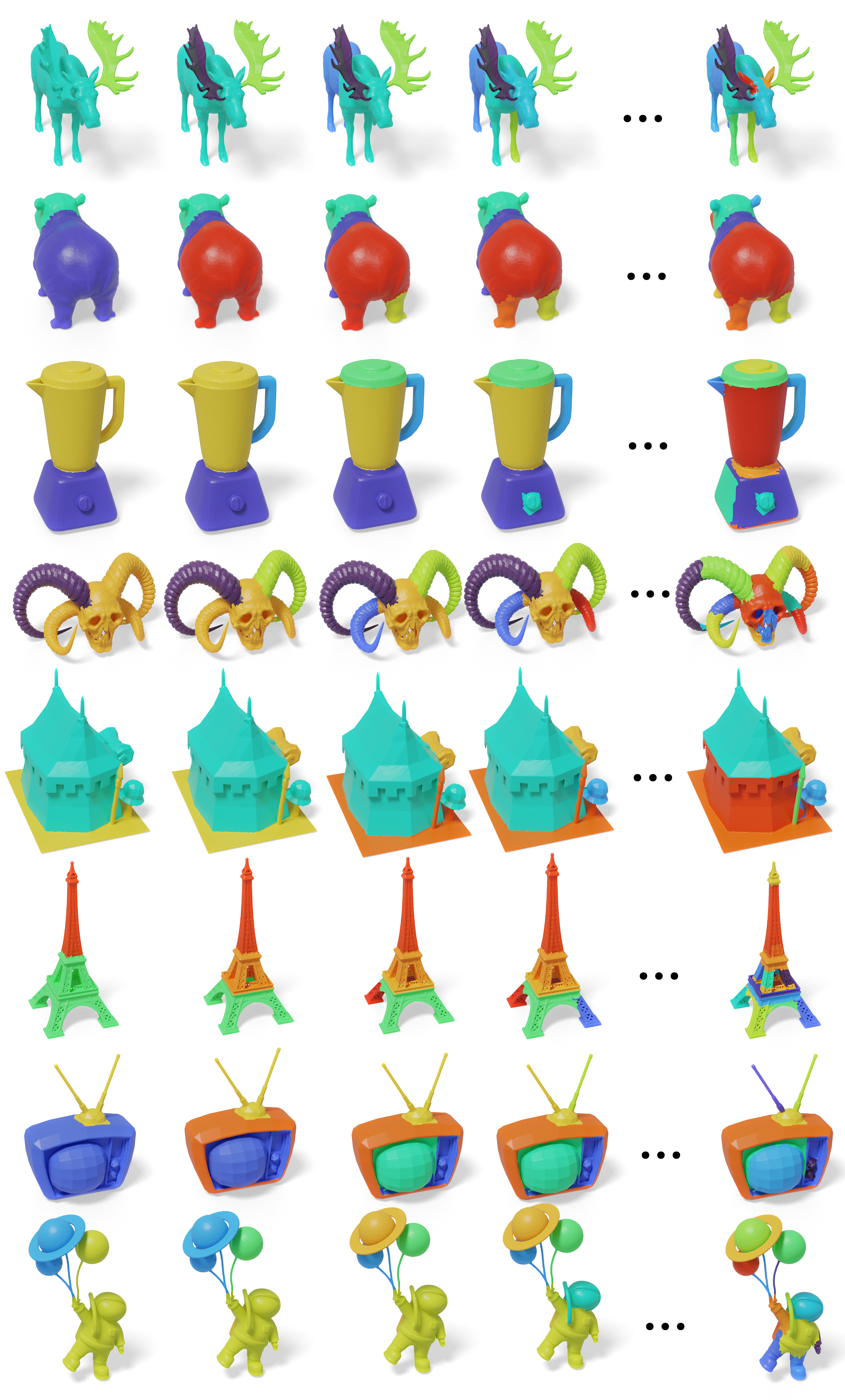}
  \caption{Additional examples of hierarchical segmentation using our method.}
  \label{fig:hierarchy_supp}
\end{figure*}

\begin{figure*}
  \centering
  \includegraphics[width=0.95\linewidth]{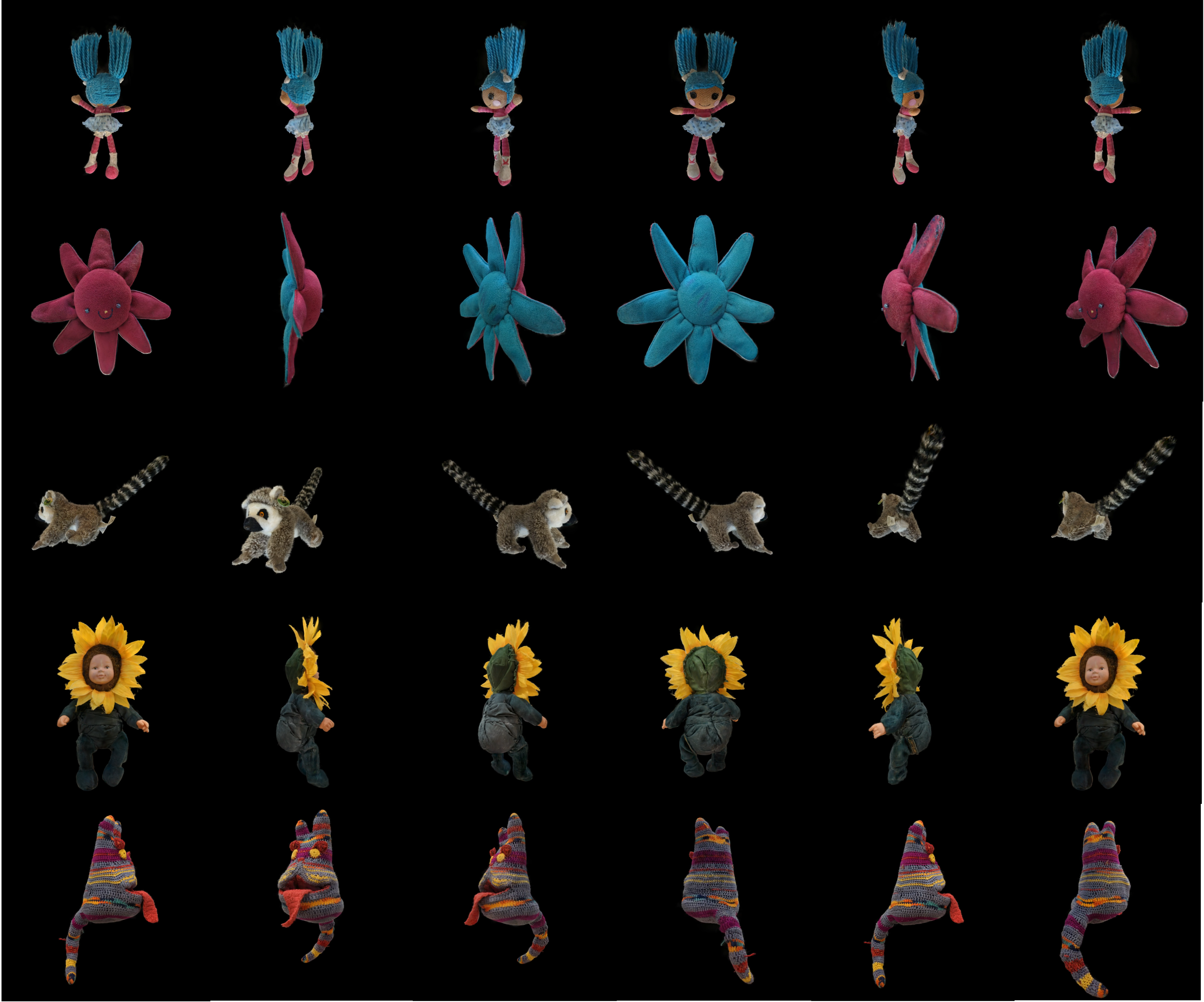}
  \caption{Sample images used for 3D Gaussian splatting reconstruction.}
  \label{fig:3dgs_images_supp}
\end{figure*}

\section{PartNetE Class Grouping}

When reporting PartNetE results, we grouped the original 45 classes into five clusters to save space. Here is the mapping:

\begin{itemize}
    \item Electronics \& Computing Devices: Keyboard, Mouse, Laptop, Phone, Camera, USB, Display (monitor), Remote, Printer, Switch (if treated as a network or power switch)
    \item Large Home Appliances: WashingMachine, Dishwasher, Refrigerator, Oven, Microwave
    \item Kitchen \& Food-Related Items: KitchenPot, Kettle, Toaster, CoffeeMachine, Faucet, Dispenser, Knife, Bottle, Bucket (often used in kitchen/cleaning contexts)
    \item Furniture \& Household Infrastructure: Table, Chair, FoldingChair, StorageFurniture, Door, Window, Lamp, TrashCan, Safe (often a household or office fixture)
    \item Tools, Office Supplies, \& Miscellaneous: Stapler, Scissors, Pen, Pliers, Lighter, Box, Cart (e.g., utility cart), Globe (decorative/educational), Suitcase (travel/personal), Eyeglasses (personal), Clock
\end{itemize}

\end{document}